\pdfoutput=1

\documentclass[11pt]{article}

\usepackage[]{eacl2023}

\usepackage{times}
\usepackage{latexsym}
\usepackage{url}
\usepackage{booktabs}

\usepackage[T1]{fontenc}

\usepackage[latin1]{inputenc}

\usepackage{microtype}

\usepackage{graphicx}
\DeclareGraphicsExtensions{.pdf,.ai,.jpg,.png}
\setkeys{Gin}{pagebox=artbox}
\graphicspath{{./eacl23-image-captions-figures/}}

\usepackage{latexsym}
\usepackage{amsmath}
\usepackage{graphicx}
\usepackage{colortbl}
\usepackage{multirow}
\usepackage{arydshln}
\usepackage{ragged2e}

\usepackage{xcolor}
\definecolor{ultralightgray}{gray}{0.90}
\definecolor{lightgray}{gray}{0.80}
\definecolor{mediumgray}{gray}{0.60}
\definecolor{darkgray}{gray}{0.40}

\newcommand{\Ni}{(1)~}
\newcommand{\Nii}{(2)~}
\newcommand{\Niii}{(3)~}

\newcommand{\mgcaption}[1]{%
  ``\textit{#1}''}

\usepackage{xspace}
\newcommand{\webisWikiICP}{Wikipedia-IPC\xspace}

\begin{document}

\title{Paraphrase Acquisition from Image Captions}

\author{
Marcel Gohsen \\
Bauhaus-Universität Weimar 
\And
Matthias Hagen\\
Friedrich-Schiller-Universität Jena \\
\AND
Martin Potthast \\
Leipzig University and ScaDS.AI \\
\And
Benno Stein \\
Bauhaus-Universität Weimar \\
}

\maketitle

\begin{abstract}
We propose to use image captions from the Web as a previously underutilized resource for paraphrases (i.e.,~texts with the same ``message'') and to create and analyze a corresponding dataset. When an image is reused on the Web, an original caption is often assigned. We hypothesize that different captions for the same image naturally form a set of mutual paraphrases. To demonstrate the suitability of this idea, we analyze captions in the English Wikipedia, where editors frequently relabel the same image for different articles. The paper introduces the underlying mining technology, the resulting Wikipedia-IPC dataset, and compares known paraphrase corpora with respect to their {\em syntactic and semantic paraphrase similarity} to our new resource. In this context, we introduce characteristic maps along the two similarity dimensions to identify the style of paraphrases coming from different sources. An annotation study demonstrates the high reliability of the algorithmically determined characteristic maps.
\end{abstract}

\section{Introduction}

Two texts that convey ``semantically equivalent information'' via a different wording are called {\em paraphrases}, or said to {\em be in a paraphrase relation}. A variety of natural language processing tasks have been approached by using paraphrase resources or paraphrase generation and detection algorithms: textual entailment \cite{dagan:2013,marelli:2014,izadinia:2015}, semantic similarity \cite{agirre:2015,li:2016}, machine translation \cite{seraj:2015}, text reformatting \cite{stein:2014a}, or question answering \cite{fader:2013}. Paraphrasing can help to improve sentence compression and text summarization \cite{cordeiro:2007a} as well as natural language understanding \cite{ganitkevitch:2013b}. 

To tackle the task of paraphrasing, large amounts of paraphrases usually are helpful. Hence, paraphrase acquisition has gained popularity over the recent years and paraphrase datasets have been created manually, \mbox{(semi-)}automatically, and via distant supervision resulting in different levels of ``naturalness'', diversity (topic, genre, language, etc.), granularity (word, sentence, or passage level), and scale (see~Section~\ref{related-work} for details). For model training, \citet{zhang:2019} strongly emphasize the importance of negative examples that are lexically similar but semantically inappropriate as paraphrases.

\begin{figure}[t]
\setlength{\fboxsep}{2pt}
\setlength{\fboxrule}{0.75pt}

\begin{tabular}[t]{@{}c@{}}
\fcolorbox{lightgray}{white}{\includegraphics[width=.2\textwidth]{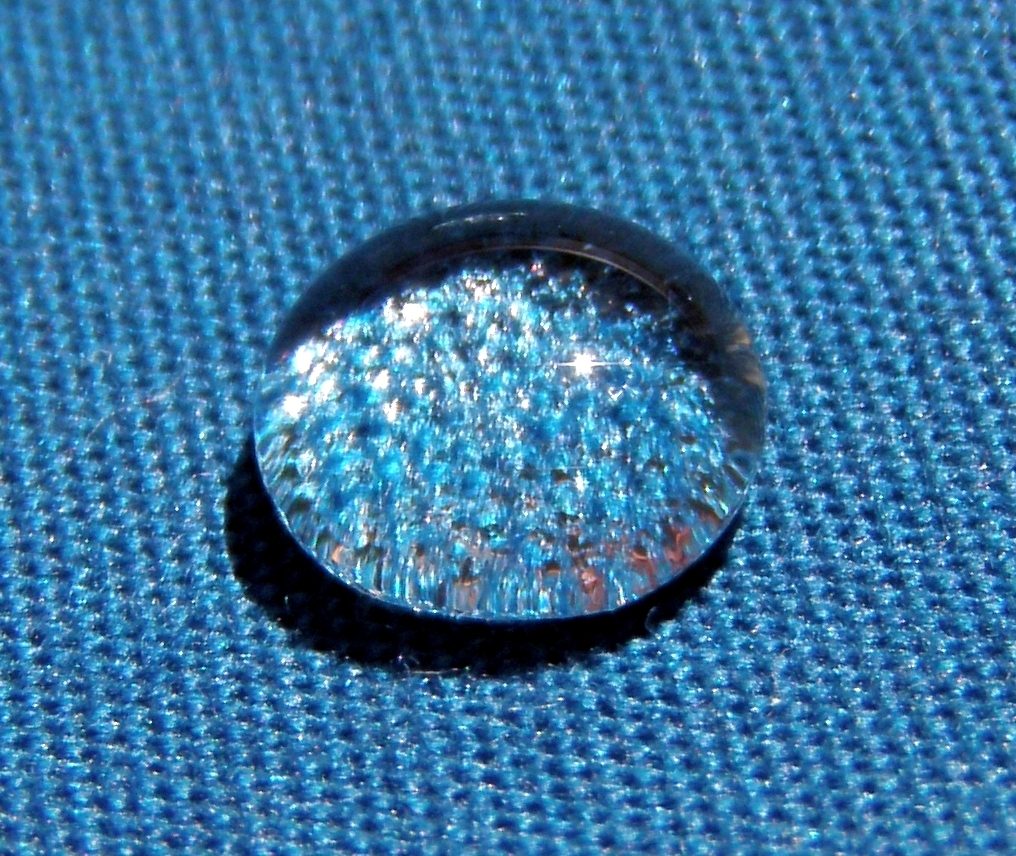}} \\[-1ex]
\parbox[t]{.22\textwidth}{\Centering\fontsize{8}{8}\selectfont\color{darkgray} Flourine-containing durable water repellent makes a fabric water-\\ resistant.}
\end{tabular}
\quad
\begin{tabular}[t]{@{}c@{}}
\fcolorbox{lightgray}{white}{\includegraphics[width=.2\textwidth]{class-c_wetting}} \\[-1ex]
\parbox[t]{.22\textwidth}{\Centering\fontsize{8}{8}\selectfont\color{darkgray} Water bead on a fabric that has been made non-wetting by chemical treatment.}
\end{tabular}
\caption{Paraphrased captions of an image in the Wikipedia pages `Durable water repellent' (left) and `Wetting' (right).}
\label{paraphrased-caption-example}
\vspace{-1em}
\end{figure}

\enlargethispage{\baselineskip}
With image captions, we propose an as yet underutilized source of natural paraphrases. The underlying hypothesis is that different captions for the same image form paraphrase candidates because the captions may describe the same image content. We study the phenomenon of `image reuse and captioning' in the English Wikipedia (Figure~\ref{paraphrased-caption-example} shows an example) by mining captions of images that have been reused. Carefully optimized filtering heuristics for captions and associated images yield a large set of captions, which we analyze qualitatively and quantitatively for their usefulness as paraphrases. The set of mined image captions forms our \webisWikiICP dataset,%
\footnote{Data: \url{https://doi.org/10.5281/zenodo.7621320}}%
\textsuperscript{,}%
\footnote{Code: \url{https://github.com/webis-de/EACL-23}}
which contains caption pairs as paraphrases along with their image source three different quality levels; 30,237~caption pairs are gold-, 229,877 are silver-, and 656,560 are bronze-quality paraphrase candidates.


\newcommand{\bscell}[2]{%
  \parbox[t]{#1\textwidth}{\raggedright #2\rule[-1ex]{0em}{0ex}}}

\newcommand{\bscellA}[2][0.12]{\bscell{#1}{#2}}
\newcommand{\bscellB}[2][0.40]{\bscell{#1}{#2}}
\newcommand{\bscellC}[2][0.40]{\bscell{#1}{#2}}

\definecolor{bleudefrance}{rgb}{0.19, 0.55, 0.91}

\newlength{\bsfigheight}

\newsavebox\imagebox
\newcommand{\bsfig}[3][1.2]{%
  \sbox{\imagebox}{\includegraphics[height=\bsfigheight]{#2}}
  \setlength{\fboxsep}{1pt}%
  \setlength{\fboxrule}{0.5pt}%
  \begin{tabular}[t]{@{}c@{}}
    \mbox{} \\[-2ex]
    \fcolorbox{lightgray}{white}{\usebox{\imagebox}} \\[-1.5ex]
    \ifx\\#3\\\else
      ~\parbox[t]{#1\wd\imagebox}{\Centering \fontsize{6}{6}\selectfont\color{darkgray} #3}\\ \fi 
    \addlinespace[2pt]
  \end{tabular}}

\begin{table*}
\small
\centering
\renewcommand{\tabcolsep}{2pt}
\renewcommand{\arraystretch}{1.0}

\begin{tabular}{@{} ccc @{}}

\toprule
\multicolumn{3}{c}{\fontsize{8}{8}\selectfont\color{black} Image content} \\
\arrayrulecolor{lightgray}
\cmidrule{1-3}
\arrayrulecolor{black}

\raisebox{-1ex}[0pt][0pt]{\bscellA{\centering\bfseries Icons, \\ symbols, \\ pictograms}}
& \multicolumn{2}{c}{\bfseries Pictures, graphics, paintings, maps, schematics} \\ 

\cmidrule{2-3}
& \multicolumn{2}{c}{\fontsize{8}{8}\selectfont\color{black} Image purpose} \\
\arrayrulecolor{lightgray}
\cmidrule{2-3}
\arrayrulecolor{black}
& \bscellB{\centering\fontsize{8}{8}\selectfont {\bfseries Ideological} \ \color{black} (used to convey atmosphere and sentiment)}
& \bscellC{\centering\fontsize{8}{8}\selectfont {\bfseries Specific, explanatory, pragmatic} \ \color{black} (used in discourse)} \\ 

\midrule[\heavyrulewidth]
\addlinespace[2ex]
\bscellA{
\setlength{\bsfigheight}{4ex}
\bsfig[1.0]{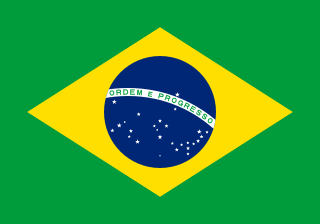}{Brazilian flag}
\setlength{\bsfigheight}{4ex}
\bsfig{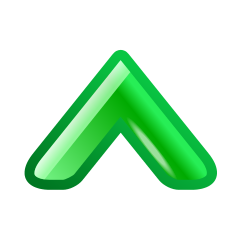}{Up arrow}

\setlength{\bsfigheight}{4ex}
\bsfig[1.6]{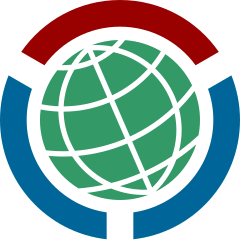}{Wikimedia community logo}
\setlength{\bsfigheight}{4ex}
\bsfig[1.6]{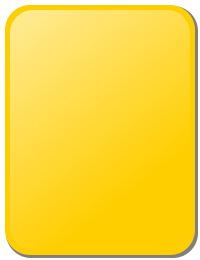}{Yellow card}

\setlength{\bsfigheight}{4ex}
\bsfig[1.5]{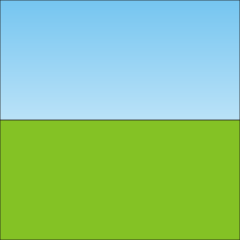}{Icon for plain stage in cycling stage race}
\bsfig[1.1]{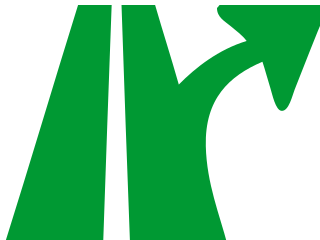}{Motorway exit icon}

\setlength{\bsfigheight}{4ex}
\bsfig[2.0]{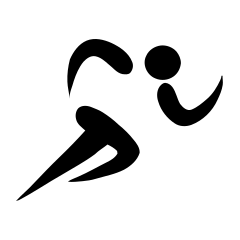}{Pictograms of Olympic sports Athletics}
\setlength{\bsfigheight}{3.0ex}
\bsfig{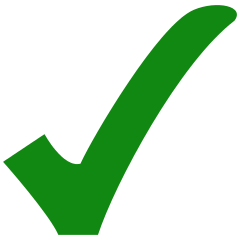}{}
}
~~
{\color{mediumgray} \vline}
\,
&

\bscellB{
\setlength{\bsfigheight}{11ex}
\bsfig{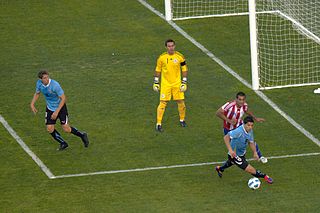}{Paraguayan goalkeeper Justo Villar was awarded as the best goalkeeper of the tournament.}
\bsfig[1.4]{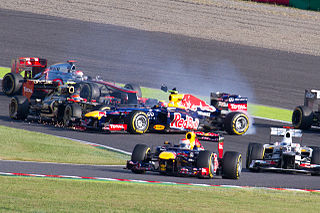}{Romain Grosjean was once again the center of controversy when he collided on the first lap with Mark Webber.}

\setlength{\bsfigheight}{7ex}
\bsfig[1.1]{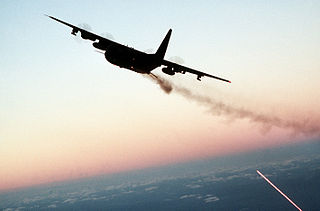}{AC-130 Spectres were highly effective during the battle.}
\bsfig[1.6]{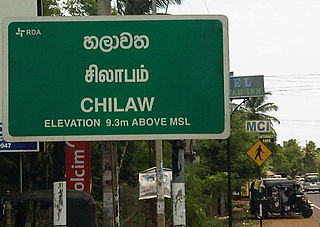}{J.\ C.\ A.\ Corea was born in the historic town of Chilaw, on the west coast of Sri Lanka.}
\bsfig{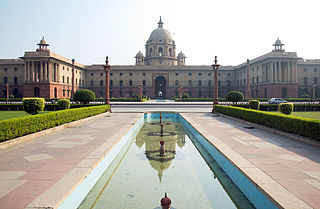}{February 10: New Delhi becomes India's capital.}

\bsfig{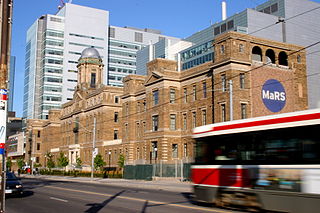}{Toronto's MaRS Discovery District is a centre for research in biomedicine.}
\bsfig[1.1]{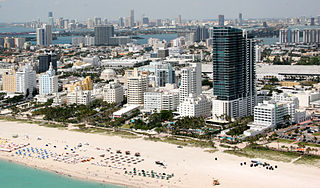}{Dostana was filmed almost entirely in Miami.}
\bsfig[1.3]{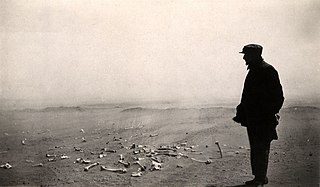}{The Armenian genocide (pictured) was the first event officially condemned as \"{}crimes against humanity\"{}.}

\setlength{\bsfigheight}{11ex}
\bsfig[2.1]{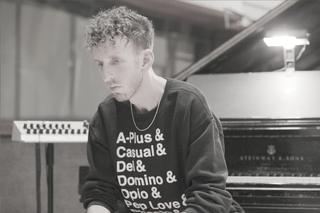}{``{}I Blame Myself'' was co-written by Ariel Rechtshaid (pictured), who additionally assisted with the songwriting and production of each track on Night Time, My Time.}
}
~
{\color{mediumgray} \vline}
~
&

\bscellC{
\setlength{\bsfigheight}{9ex}
\bsfig[2.0]{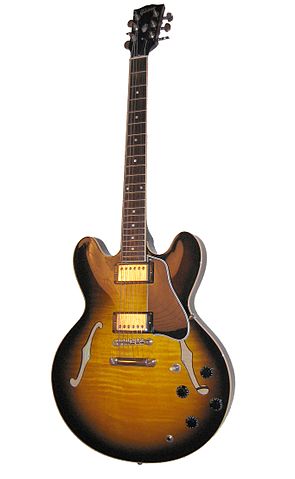}{Semi hollow-body electric guitar Gibson ES-335 has a \"{}solid center block\"{} inside a body.}
\bsfig[1.7]{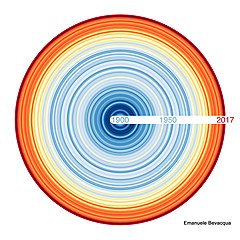}{This \"{}circular warming stripes\"{} graphic depicts average global temperature using chronologically ordered, concentric coloured rings.}
\bsfig{class-c_wetting}{
 Water bead on a fabric that has been made non-wetting by chemical treatment.
}

\setlength{\bsfigheight}{9ex}
\bsfig[1.1]{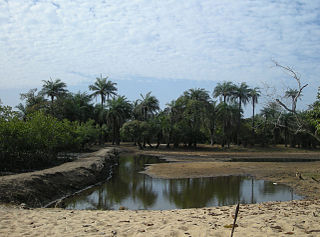}{Dikes are used to protect the rice paddy fields from the channels of saltwater which overflow during high tide.}
\setlength{\bsfigheight}{7ex}
\bsfig[1.1]{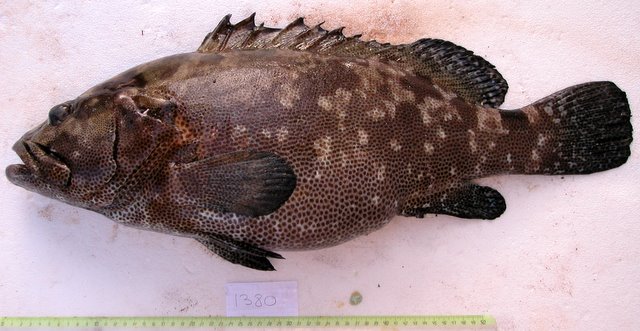}{The camouflage grouper Epinephelus polyphekadion is the type-host of P.~viscosus; the photograph shows a specimen from New Caledonia.}
\bsfig[1.1]{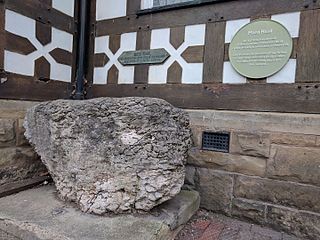}{The stone is against the half-timbered wall of Exmewe Hall, on St.~Peter's Square, Ruthin.}

\bsfig[1.8]{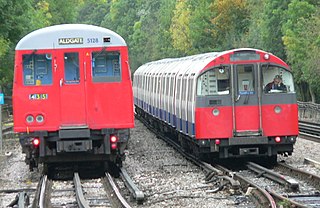}{A sub-surface Metropolitan line A Stock train (left) passes a deep-tube Piccadilly line 1973 Stock train (right) in the siding at Rayners Lane.}
\bsfig[2.4]{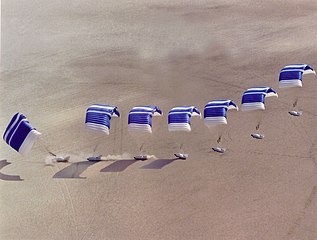}{The X-38 CRV prototype makes a gentle lakebed landing at the end of a July 1999 test flight at the Dryden Flight Research Center with a fully deployed parafoil.}

\bsfig[1.0]{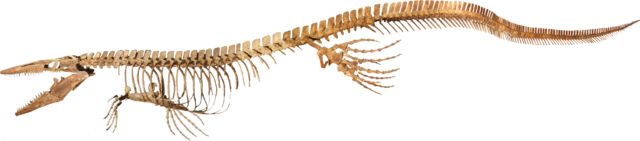}{Tylosaurus proriger mounted skeleton in the Rocky Mountain Dinosaur Resource Center in Woodland Park, Colorado.}
}
\\
\addlinespace[1ex]

\bottomrule
\end{tabular}
\caption{Example Wikipedia images and captions. Following \citet{baer:2008}, we categorize images by content as icons, symbols, pictograms (left), or pictures, graphics, etc.\ (middle and right). Pictures, graphics, etc.\ can be further subdivided by the purpose they serve in a text as either ideological (i.e., to convey sentiment; middle) or pragmatic (i.e., to explain and specify; right).}
\label{table-image-and-caption-examples}
\end{table*}

The usefulness of an image caption as a paraphrase depends strongly on its corresponding image. Table~\ref{table-image-and-caption-examples} shows example image--caption pairs from Wikipedia, organized by content and purpose classes. Icons, symbols, and pictograms rarely have captions that are useful as paraphrase candidates, while for pictures, graphics, paintings, maps, and schematics, the purpose of the image in a page is crucial. In this regard, ``ideological purpose'' means that the image is mainly used for aesthetic reasons, resulting in captions that do not describe the shown content but rather some context, atmosphere, or sentiment. By contrast, explanatory or pragmatic images try to illustrate or transport knowledge. Their captions describe and explain (parts of) the visual content and form a particularly promising source of natural paraphrases.

\section{Related Work}
\label{related-work}

\paragraph{Paraphrase Corpora}

To collect high-quality paraphrases, some studies applied manual acquisition. For instance, for the Webis-CPC-11 \cite{burrows:2013}, crowdworkers were asked to rewrite one of 4,096~pass\-age-level literature excerpts \textit{``so that the rewritten version has the same meaning, but a completely different wording and phrasing.''} The resulting 7,859~pairs form one of the few passage-level paraphrase corpora. Later, also \citet{xu:2014} crowdsourced sentences with the same meaning as a shown sentence from some tweet for their PIT-2015 dataset (18,862~sentence pairs, 5,641~considered paraphrases).

Since crowdsourcing is costly to scale, automatic paraphrase acquisition often simply relies on machine translation. For instance, \citet{zhang:2019} scrambled words and forth-and-back translated sentences to obtain a gold dataset of 108,463~paraphrase pairs and a silver dataset of 656,000~pairs. Also, the ParaNMT-50M dataset \cite{wieting:2018} with 30~million ``strong'' paraphrases (assessment by the authors), was created by translating sentences from English-Czech parallel pairs.

Trying to combine the advantages of manual and automatic methods (quality vs.\ scale), most paraphrase acquisition methods use distant supervision. For instance, \citet{barzilay:2001} suggested aligned sentences from monolingual parallel corpora as paraphrases---an idea later used by \citet{dolan:2005} to create the Microsoft Research Paraphrase Corpus (5,801~manually annotated sentence pairs, 3,900~considered paraphrases). Instead of monolingual sources, \citet{bannard:2005} exploited bilingual parallel corpora by aligning phrases that translate into the same pivot phrase in another language. Based on this idea, the English/Spanish PPDB~1.0 \cite{ganitkevitch:2013a} and its extension to 23~additional languages \cite{ganitkevitch:2014} with an English portion of 7.6~million lexical (i.e., synonymous) and 68.4~million phrasal paraphrases form the biggest paraphrase collection.

Translation pivoting was also used for the Opusparcus \cite{creutz:2018} and TaPaCo \cite{scherrer:2020} datasets. Opusparcus was created from movie and TV subtitles in six European languages. The automatically filtered training set contains 7~million English sentence and phrase pairs (``good'' or ``mostly good'' paraphrases according to the authors) while the manually labeled development and test sets contain 3,088~pairs (1,997~paraphrases). Analogously, TaPaCo is created from the multilingual Tatoeba data, a crowdsourced collection of sentence translations.%
\footnote{\url{https://tatoeba.org/eng/about}}
The English part of TaPaCo contains 158,000~sentences that are clustered in 62,000~paraphrase sets. Instead of translation, \citet{lan:2017} used a different pivoting method for their Twitter News URL Corpus. They mined tweets that contain the same hyperlink and labeled 51,524~respective sentence pairs via crowdsourcing. Our paraphrase acquisition methodology is also based on a pivoting technique, but the pivot media are images.

\paragraph{Captions and Paraphrases}

Our idea to use images from the web as pivots for distant supervision paraphrase acquisition is inspired by studies on crowdsourced caption datasets. For instance, \citet{marelli:2014} extracted a collection of 9,840~paraphrase candidates from the Flickr~8k dataset \cite{rashtchian:2010} of 8,000~images each with 5~crowdsourced captions and the MSR~Video Paraphrase Corpus \cite{chen:2011} of 2,000~short video clips with 85,000~English crowdsourced descriptions of what can be seen. The larger MSCOCO dataset \cite{lin:2014} with 120,000~images each having 5~crowdsourced captions was even already used to train a neural paraphrase generation model \cite{prakash:2016}. Interestingly, the existing caption-paraphrasing datasets were all created in dedicated crowdsourcing processes that are costly to scale. We are the first to explore mining ``real'' image captions from the web (i.e., not crowdsourced) as a paraphrase acquisition method.

\section{Caption-Based Paraphrase Acquisition}

We collect captions of images that are used on the English Wikipedia, grouping those captions that describe the same image. This section outlines the mining and filtering steps.

\paragraph{Data Source: Wikipedia}

As a source for paraphrases, we use captions from Wikipedia articles. Wikipedia is a collection of well-formulated, human-written text that is constantly being improved by a community of volunteer editors. The guidelines for the use of images on Wikipedia encourage editors to write a caption to explain the relevance of an image in a particular context, resulting in 60\% of images having a caption.

We use the English Wikipedia for the paraphrase acquisition (Wikimedia, September~1st,~2022).%
\footnote{\url{https://dumps.wikimedia.org/enwiki/}}
Two versions of Wikipedia dumps are available for the paraphrase acquisition---a reduced dataset with all articles, templates, media descriptions, and primary meta pages and a full-size dump with all pages, including their full revision history. Table~\ref{table-wikimedia-dumps-stats} compares the number of pages, revisions, images, and image references between these two datasets.

\begin{table}

\footnotesize
\centering
\setlength{\tabcolsep}{2pt}
\renewcommand{\arraystretch}{1.2}

\begin{tabular}{@{}l rrrr@{}}
\toprule
  \bfseries Dump & \multicolumn{1}{@{}c@{}}{\bfseries Pages} & \multicolumn{1}{@{}c@{}}{\bfseries Revisions} & \multicolumn{1}{@{}c@{}}{\bfseries Images} & \multicolumn{1}{@{}c@{}}{\bfseries Image refs.} \\
\midrule
  Small          & 22,311,116                                &                                    22,311,116 &                                  4,508,808 &                                       7,384,368 \\
  Full           & 54,898,564                                &                                   987,527,838 &                                 11,183,853 &                                   3,629,447,851 \\
\bottomrule
\end{tabular}
\caption{Image mining statistics of Wikipedia dumps. \mgcaption{Small} refers to all articles, templates, media descriptions, and primary meta-pages without edit histories, whereas \mgcaption{Full} includes edit histories.}
\label{table-wikimedia-dumps-stats}
\vspace{-1em}
\end{table}

\begin{table*}

\fontsize{8pt}{9pt}\selectfont
\centering
\setlength{\tabcolsep}{3.6pt}
\renewcommand{\arraystretch}{1.0}

\begin{tabular}{@{}l@{} rrrr@{\quad} rrr@{\quad} rr@{\quad} rr@{}}
\toprule
                              &                       \multicolumn{4}{c}{\textbf{Images}}                       &        \multicolumn{3}{c}{\textbf{Image references\quad\ }}         & \multicolumn{2}{c}{\textbf{Image captions\quad\ }} &  \multicolumn{2}{c}{\textbf{Paraphrase cand.}}   \\
\cmidrule(l{0pt}r{3\tabcolsep}){2-5}\cmidrule(r{3\tabcolsep}){6-8}\cmidrule(r{3\tabcolsep}){9-10}\cmidrule{11-12}
  \textbf{Filter}             & Remaining  & \multicolumn{1}{@{}c@{}}{$\Delta$} & Refs. $\geq2$ & Refs. $\geq5$ & Remaining & \multicolumn{1}{@{}c@{}}{$\Delta$} & $\mu_{\text{img}}$ & Remaining &     \multicolumn{1}{@{}c@{}}{$\Delta$} & Remaining   & \multicolumn{1}{@{}c@{}}{$\Delta$} \\[0.5ex]
\midrule\addlinespace
  0. No filter                & 4,,508,808 &       \multicolumn{1}{@{}c@{}}{--} &       941,339 &       102,046 & 7,384,368 &       \multicolumn{1}{@{}c@{}}{--} &               1.64 & 4,571,671 &           \multicolumn{1}{@{}c@{}}{--} & 355,619,088 &       \multicolumn{1}{@{}c@{}}{--} \\[1ex]
\hdashline\addlinespace
  1. References $\geq$ 2      & 941,339    &                              -79\% &       941,339 &       102,046 & 3,816,899 &                              -48\% &               4.05 & 2,053,338 &                                  -55\% & 355,619,088 &                                0\% \\
  2. References $\leq$ 10     & 918,758    &                               -2\% &       918,758 &        79,465 & 2,494,271 &                              -35\% &               2.71 & 1,569,327 &                                  -23\% & 1,350,050   &                             -100\% \\
\addlinespace
  3. Has caption              & 713,815    &                              -22\% &       494,321 &        32,949 & 1,499,479 &                              -40\% &               2.10 & 1,569,327 &                                    0\% & 1,350,050   &                                0\% \\
  4. Caption words $\geq$ 6   & 518,851    &                              -27\% &       300,244 &        18,436 & 987,541   &                              -34\% &               1.90 & 1,014,351 &                                  -35\% & 760,448     &                              -44\% \\
  5. Caption is sentence      & 97,496     &                              -81\% &        23,062 &           665 & 128,893   &                              -87\% &               1.32 & 129,776   &                                  -87\% & 44,005      &                              -94\% \\
\addlinespace\hdashline\addlinespace
  6. References $\geq$ 2      & 23,062     &                              -76\% &        23,062 &           665 & 54,459    &                              -58\% &               2.36 & 54,940    &                                  -58\% & 44,005      &                                0\% \\
  7. Unique candidates        & 22,643     &                               -2\% &        22,582 &           507 & 52,394    &                               -4\% &               2.31 & 52,394    &                                   -5\% & 39,995      &                               -9\% \\
  8. Divergent captions       & 18,979     &                              -16\% &        18,979 &           391 & 43,616    &                              -17\% &               2.30 & 43,616    &                                  -17\% & 32,830      &                              -18\% \\
  9. Significant caption diff & 17,293     &                               -9\% &        17,293 &           386 & 40,131    &                               -8\% &               2.32 & 40,131    &                                   -8\% & 30,237      &                               -6\% \\
\bottomrule
\end{tabular}
\caption{Effects of the filtering steps in the paraphrase acquisition pipeline on the number of images, references, captions, and paraphrase candidates mined from the reduced Wikipedia dump. Images that have been referenced at least 2 or 5 times are called ``Refs. $\geq2$'' and ``Refs. $\geq5$'', respectively. $\mu_{\text{img}}$ describes the average number of references per img.}
\label{table-caption-filter-application}
\end{table*}

\paragraph{Image and Caption Mining}

Wikimedia Commons is a service which hosts free-to-use images and other media files which are used across all Wikimedia projects (like Wikipedia). The markup language for Wikipedia articles, Wikitext, features the \textit{extended image syntax}, which allows editors to insert images into Wikipedia pages from Wikimedia Commons. The specified image is referenced by its unique identifier in Wikimedia Commons, which is a concatenation of a generic media type prefix (e.g., Image) followed by a colon and its (also unique) filename. The extended image syntax allows the specification of style properties, meta-information, and a caption. If a caption is present, we store it along the image reference.

However, not all images in Wikipedia pages are included by the use of the extended image syntax. Wikitext allows the use of (user-created) style templates, often with inconsistent image reference syntax. \textit{Infoboxes} on Wikipedia are template-based tables placed at the beginning of an article that contain a collection of the most important information about its subject. They are commonly used in Wikipedia---20\%~of all the pages contain one. Important to us is that one third of the infoboxes contain one or even multiple images, and about~56\% of these images have a caption. Wikitext represents infoboxes as lists of key-value pairs whose sets of allowed keys and values are defined in templates. There are close to 2,000~different infobox templates based on semantic categories of an article. The responsible keyword for image referencing varies among these templates, and to cover all cases is not feasible. Thus, we identified the most commonly used keyword which is ``image'', followed by an optional counter for multiple images. These captions are mined as individual image--caption pairs. No Wikitext parser other than Wikipedia's own implementation (which is monolithically intertwined with Wikipedia's software stack) is capable of reliably parsing all the different template syntaxes. Our heuristic for image mining from infoboxes yields an increase of~24\% of image references.

In addition to a caption, an image reference may also have an \textit{alternative text} that accurately describes the visual content of an image.  Alternative texts are intended to serve as a substitute in case the user is visually impaired, or the image cannot be displayed correctly. According to Wikipedia's policy, all images are obliged to have an alternative text, except those with a purely decorative purpose. We extract them as paraphrase candidates, too. Altogether, 355~million caption pairs are mined that we subsequently filter as follows (see Table~\ref{table-caption-filter-application}).

\paragraph{Image Filter}

A necessary condition for an image to be useful is that it has been referenced at least twice, so we discard all image captions whose images have only one reference.  As illustrated in Table~\ref{table-image-and-caption-examples}, the likelihood of image--caption pairs to form a pair of paraphrase candidates depends on the image's content and purpose. A useful feature to discard icons, symbols, and pictograms is the number of references. These images occur more frequently than any other kind of image. To discard them, we set an upper bound of 10~references per image for the corresponding image--caption pairs to be retained. After these steps, 1.35 million image--caption pairs remain (see Table~\ref{table-caption-filter-application}).

\paragraph{Caption Filter}

The selection of caption pairs has a substantial effect on the quality of the acquired paraphrases. As a consequence, all mined image captions pass through various carefully optimized pre-processing and filtering steps. The pre-processing pipeline comprises cleansing of Wikitext markup, removal of line-breaks, and white-space normalization (e.g., removal of sequences).

To discard trivial image captions, we determined a lower bound for the caption length through manual observation and found that most captions that represent a phrasal expression contain at least 6~words. Captions with less than that usually comprise the name of an entity that can be seen in the image and are consequently discarded.

Through manual review, we identified three common practices for writing image captions with respect to style and syntax. Captions on Wikipedia are either:
\Ni
simple noun phrases (e.g., \mgcaption{Last Supper by Dieric Bouts}),
\Nii
sentence fragments containing a verb phrase (e.g., \mgcaption{Last Supper drawn by Dieric Bouts}), or
\Niii
grammatically correct sentences (e.g., \mgcaption{The Last Supper was drawn by Dieric Bouts}).
The latter two caption types are more likely to be a source of high-quality paraphrases, whereas grammatical sentences are most desirable. To discard noun phrases, it is sufficient to exclude captions that do not contain verbs. The detection of proper sentences is far more difficult.

Grammaticality and sentence fragment detection is a broad research field and most state-of-the-art methods use neural models. However, to ensure high precision in the resulting paraphrase dataset, we resorted to a more efficient and effective heuristic that is well suited for captions: rule-based sentence classification. Since the authors of existing rule-based grammar checkers rarely publish their full rule sets, we create our own rules to identify grammatical sentences. To find suitable rules, we take 500 and 100 random captions from Wikipedia to create a training and test set, respectively. An expert with 20 years of experience in the field of linguistics manually annotated the 600~captions to determine whether they are sentences or sentence fragments. Based on the 500~annotations from the training set and the associated automatically assigned POS tags, the expert manually created a set of rules to distinguish fragments from sentences.

\begin{table}[t]
\fontsize{9pt}{10pt}\selectfont
\centering
\setlength{\tabcolsep}{3pt}
\renewcommand{\arraystretch}{1.2}

\begin{tabular}{@{}c ll@{}}
\toprule
  \textbf{Rule} & \textbf{Premise}               & \textbf{Pattern}                                 \\
\midrule
        1       & $.^*\texttt{MD}.^*$            & $.^*\ \texttt{MD RB? VB}\ .^*$                   \\
        2       & $.^*\texttt{(WDT|WP|WRB)}.^* $ & $[\neg(\texttt{WDT|WP|WRB})]^*$                  \\[-0.5ex]
                &                                & $\texttt{(VBP|VBZ|VBD)}.^*$                      \\
        3       & $.^*\texttt{IN}.^*$            & $[\neg \texttt{IN}]^* \texttt{(VBP|VBZ|VBD)}.^*$ \\
        4       & $\bot$                         & $.^*\texttt{(VBP|VBZ|VBD)}.^*$                   \\
\bottomrule
\end{tabular}
\caption{POS-based rules for sentence classification.}
\label{table-sentence-patterns}

\end{table}

Table~\ref{table-sentence-patterns} shows our sentence classification rules based on POS~tags from Penn Treebank~\cite{taylor:2003}. To be considered a sentence, a caption must satisfy one of these rules. Which rule should be applied is determined by its premises, which are tested in the order given. The first rule targets sentences that contain modals. A modal must be directly followed by an optional adverb and an obligatory verb in the base form. A sentence that would be correctly classified as such is \mgcaption{Last Supper might be drawn by Dieric Bouts}. The second and third rules deal with subordinate clauses at the end of a sentence. These rules state that an inflected verb must precede a subordinating conjunction. For example, \mgcaption{Last Supper was drawn by Dieric Bouts which is an exceptional artwork} would be correctly classified as a sentence under Rule~2. Rule 4 requires that sentences contain an inflected verb and is applied when no other premise is satisfied. \mgcaption{Dieric Bouts drew the Last Supper} is a correctly classified sentence according to Rule~4.

Our heuristic sentence classifier performs satisfactorily with a precision of ~94\% and a recall of ~79\%, evaluated on the 100~annotated captions in the test set. Problematic cases are verbs that have identical base and inflected forms, which are often mislabeled by the POS tagger of the Stanford CoreNLP toolkit \cite{manning:2014}.

After application of these caption filters, in total 44,005 image--caption pairs remain (see Table~\ref{table-caption-filter-application}).

\paragraph{Image Equivalence}

Wikipedia hosts all of its media files on Wikimedia Commons, and therefore we can identify duplicate images without requiring physical copies of an image. All images have a unique name that, along with a file type prefix, points to the URI of their description page on Wikimedia Commons. When an editor of a Wikipedia article wants to include an image from Wikimedia Commons, the image is referenced by this URI. Therefore, this URI is the criterion for deciding whether images are equivalent. All file type prefixes are derived from the common ``File'' prefix. Therefore, all image prefixes are normalized to ``File'' prefixes to detect image duplicates with inconsistent prefix usage.

\paragraph{Paraphrase Construction}

All references that refer to the same image are grouped into clusters. Paraphrase candidates are constructed by forming all possible unique pairs of captions of the same caption type (i.e., alternative text or regular caption) within each cluster. Combining regular captions and alternative texts rarely results in meaningful paraphrases due to their different purposes. Alternative texts describe the visual content of an image, while regular captions describe the image in context and from the perspective of an article.

Of these paraphrase candidates, we discard caption pairs that are exact or near-duplicates. We consider a caption pair to be a near-duplicate if it differs only by punctuation, capitalization, or whitespaces. (Near-)duplicates of captions are artifacts of reusing centrally hosted captions for the same image in different articles on Wikipedia. Removing duplicates results in 30,237 caption pairs as paraphrase candidates (see Table~\ref{table-caption-filter-application}).

\paragraph{Corpus Construction}

We create three individual paraphrase datasets with different quality levels which we assign based on the used Wikipedia dump and the set of applied filtering heuristics.

Gold-quality paraphrase candidates are pairs of acquired image captions from the reduced Wikipedia dump which are classified as (sequences of) sentences by our rule-based classifier. Table~\ref{table-caption-filter-application} shows the loss of images, references, captions, and paraphrase candidates due to the application of our filter heuristics. In total, 30,237~gold-quality paraphrase candidates were acquired with our pipeline.

The most ``aggressive'' filter heuristic in the pipeline is the sentence classification of image captions. It discards~87\% of the mined image captions. Since phrases can also be in a paraphrase relation, we construct a silver-quality paraphrase dataset by exchanging the sentence classifier for a heuristic that requires image captions to contain a verb. This heuristic excludes 50\%~of image captions for the paraphrase acquisition. This altered pipeline mined 229,877~silver-quality paraphrase candidates.

For the bronze-quality paraphrase dataset, we collect as many reasonable caption-pairs as possible, using the full Wikipedia dump with edit history as source for the paraphrase acquisition. We set the upper bound of allowed number of references per image to~18, the average number of revisions per page. Thus, an image can be referenced at most 180~times. The verb heuristic is applied. This yields 656,560 bronze-quality paraphrase candidates.

\section{Corpus Analysis and Evaluation}

We analyze and evaluate our paraphrase corpus based on the quantification of paraphrase similarity, for which we select well-known syntactic and semantic measures, and compare our corpus with five state-of-the-art paraphrase and three image--caption corpora from the literature. Therefore, we propose a new measure $\Delta_{\text{sem},\text{syn}}$ that measures paraphrase sophistication based on the difference of semantic and syntactic similarities. To validate the proposed measure we analyze the correlation between human judgments and automatically computed semantic and syntactic similarities.

\subsection{Analyzing Paraphrase Similarity}

According to \citet{wahle:2022}, high-quality paraphrases are texts pairs with high semantic similarity and high lexical and syntactic diversity, as paraphrasing models trained on linguistically diverse examples tend to be more robust \cite{qian:2019}. Similarly, \citet{niu:2021} measures paraphrase quality by rewarding high semantic similarity while penalizing high lexical overlap. The relation between semantic and syntactic similarity characterizes the ``sophistication'' of a paraphrase.  Based on this assumption, we propose $\Delta_{\text{sem},\text{syn}}$ to quantify the sophistication of paraphrases. $\Delta_{\text{sem},\text{syn}}$ computes the average difference between the average semantic and syntactic similarity scores.

\paragraph{Similarity Measures}

\providecommand{\bsc}{\color{darkgray}}

\begin{table*}

\fontsize{8pt}{10pt}\selectfont
\centering
\setlength{\tabcolsep}{3pt}
\renewcommand{\arraystretch}{1.0}

\begin{tabular}{@{}l@{} c@{\ }c@{\ }cc@{\quad} cccc@{\ \ } c@{}}

\toprule
 & \multicolumn{4}{c}{\bf  Syntactic similarity} & \multicolumn{4}{c}{\bf Semantic similarity} &\\

\cmidrule(r{\tabcolsep}r{\tabcolsep\ \ }){2-5}
\cmidrule(r{\tabcolsep}r{\tabcolsep\ }){6-9}

\bf Image Caption Pairs & ROUGE-1 & ROUGE-L & BLEU & Avg.& WMS & BERT & ST & Avg. & $\Delta_{\text{sem}, \text{syn}}$\\
\midrule
\textbf{A1}: An Easter postcard from 1907 depicting a rabbit. & \multirow{2}{*}{0.53} & \multirow{2}{*}{0.13} & \multirow{2}{*}{0.14} & \multirow{2}{*}{0.27} & \multirow{2}{*}{0.76} & \multirow{2}{*}{0.76} & \multirow{2}{*}{0.90} & \multirow{2}{*}{0.81} & \multirow{2}{*}{\phantom{-}0.54}\\
\textbf{A2}: A 1907 postcard featuring the Easter Bunny.\\
\addlinespace

\textbf{B1}: Twelfth century illustration of a man digging. & \multirow{2}{*}{0.13} & \multirow{2}{*}{0.13} & \multirow{2}{*}{0.10} & \multirow{2}{*}{0.12} & \multirow{2}{*}{0.51} & \multirow{2}{*}{0.47} & \multirow{2}{*}{0.83} & \multirow{2}{*}{0.60} & \multirow{2}{*}{\phantom{-}0.48}\\
\textbf{B2}: An English serf at work digging, c. 1170. \\

\addlinespace
\hdashline
\addlinespace

\textbf{C1}: Troops clearing rubble after the May air raid on Belfast. &  \multirow{2}{*}{0.90} & \multirow{2}{*}{0.90} & \multirow{2}{*}{0.99} & \multirow{2}{*}{0.93} & \multirow{2}{*}{0.92} & \multirow{2}{*}{0.89} & \multirow{2}{*}{0.98} & \multirow{2}{*}{0.93} & \multirow{2}{*}{\phantom{-}0.00}\\
\textbf{C2}: Soldiers clearing rubble after the May air raid on Belfast.\\
\addlinespace

\textbf{D1}: System of a Down is composed of four Armenian-Americans. & \multirow{2}{*}{0.42} & \multirow{2}{*}{0.42} & \multirow{2}{*}{0.33} & \multirow{2}{*}{0.39} & \multirow{2}{*}{0.50} & \multirow{2}{*}{0.07} & \multirow{2}{*}{0.32} & \multirow{2}{*}{0.30} & \multirow{2}{*}{-0.09}\\
\textbf{D2}: Dolmayan drumming with System of a Down in 2011.\\
\bottomrule
\end{tabular}

\caption{Examples from the \webisWikiICP dataset with particularly high and low $\Delta_{\text{sem}, \text{syn}}$. Caption pairs with high $\Delta_{\text{sem}, \text{syn}}$ are structurally and lexically diverse while maintainig high levels of semantic similarity. Caption pairs with low $\Delta_{\text{sem}, \text{syn}}$ are either semantically dissimilar or have a high degree of lexical overlap.}\
\label{table-paraphrase-examples}
\end{table*}

To measure syntactic similarities between pairs of paraphrases, we choose ROUGE-1, ROUGE-L \cite{lin:2004}, and BLEU \cite{papineni:2002}. ROUGE-1 assesses unigram overlap, ROUGE-L computes lexical similarity based on the longest-common-subsequence paradigm, and BLEU is computed up to 4-grams to quantify structural similarity. All three measures have been proposed and evaluated for paraphrase generation and acquisition.

The computation of semantic similarity is done by transformer-based models. One of our three measures is BERTScore \cite{zhang:2020} since it was found to correlate well with human judgments for the task of automatic image captioning. The second one is the cosine similarity of dense vector representations computed by a BERT-based Sentence Transformer \cite{reimers:2019}. It proved to be decisive at Sent-Eval on the MSRPC. The third one is Word Mover Distance \cite{kusner:2015} which computes the minimum amount of distance that embedded words of a text need to ``travel'' to reach the embedded words of another text. We use the inversion of this measure as a Word Mover-based similarity score (WMS). All transformer-based models were applied with default hyperparameter configurations using a single NVIDIA~A100 Tensor Core~GPU. We normalize all the syntactic and semantic similarity measures to a scale from zero to one before we compute the average syntactic and semantic similarity. 

Table~\ref{table-paraphrase-examples} shows particularly sophisticated and unsophisticated caption pairs as paraphrases quantified by  $\Delta_{\text{sem},\text{syn}}$ from the \webisWikiICP (silver quality). Caption A1 is a paraphrase of A2 with a significantly different structure and sufficiently high lexical diversity. There is some unigram overlap, which causes the ROUGE-1 score to be comparatively high, since these captions are quite short. Due to the high semantic similarity of A1 and A2 these mutual paraphrases score high in the $\Delta_{\text{sem},\text{syn}}$ measure. B1 and B2 have almost no words in common, but they are nevertheless semantically close. One semantic difference is that an ``English serf'' is a different concept than ``a man'' is, which could be the reason for the comparatively lower semantic similarity. However, the semantic similarity is high enough to
yield a high $\Delta_{\text{sem},\text{syn}}$. In (C1, C2) and (D1, D2) we see caption pairs that do not qualify as ``good'' paraphrases for two different reasons. In C2, a single word is replaced by a synonym in C1 and therefore has almost identical wording to C1. Although D1 and D2 are captions that both refer to the band
``System of a Down'', they describe entirely different aspects, which is reflected in the semantic similarity scores. Therefore, both pairs (C1, C2) and (D1, D2) receive a low $\Delta_{\text{sem},\text{syn}}$ value.

\paragraph{Manual Similarity Judgments}

In order to investigate whether the chosen similarity measures accurately quantify syntactic and semantic similarity, we conduct a manual annotation study to measure the correlation between the measures and human similarity ratings. For that purpose, we sampled 100 paraphrases from each of the three datasets--\webisWikiICP, MSRPC, and TaPaCo. We draw paraphrase examples with a length of 10 to 30~words stratified by their $\Delta_{\text{sem},\text{syn}}$ distribution in their corresponding datasets. Four expert annotators with at least six years of experience rate semantic and syntactic similarity of all the 300 examples on a 5-point Likert scale. 

\begin{figure*}%
\small%
\includegraphics{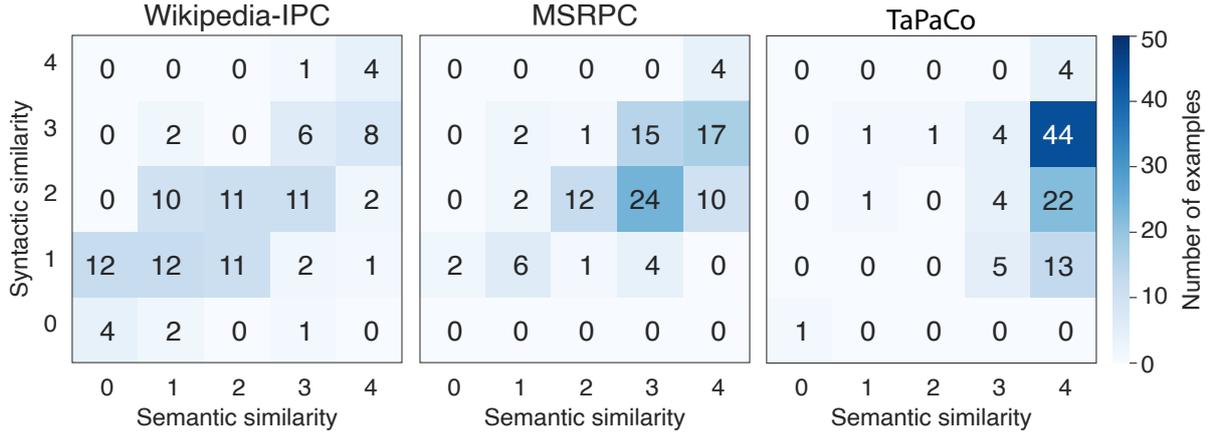}
\caption{Distribution of semantic and syntactic similarity (characteristic map) for three datasets at sample size~100 for paraphrases with a length between 10 and 30 words. Among others, it can be seen that the TaPaCo corpus comprises paraphrases that result from subtle and minor changes within a paraphrase's sentence pairs (44 x syn-4/sem-3), while Wikipedia-IPC shows a more uniform distribution along the diagonal syn-0/sem-0\ldots syn-4/sem-4.}%
\label{fig:distribution-manual-judgements}%
\end{figure*}

\begin{table}[t]

\small
\centering
\setlength{\tabcolsep}{4pt}
\renewcommand{\arraystretch}{1.2}

\begin{tabular}{@{}l ccc @{\hspace{2em}} c@{}}
\toprule
Syntax: & \bfseries \kern-1em ROUGE-1 & \bfseries ROUGE-L & \bfseries BLEU & \bfseries Average \\
\midrule
$r$ & 0.78 & 0.77 & 0.70 & 0.79 \\[2ex]
\toprule
Semantics: & \bfseries WMS & \bfseries BERT & \bfseries ST & \bfseries Average \\
\midrule
$r$ & 0.59 & 0.70 & 0.78 & 0.76 \\
\bottomrule
\end{tabular}
\caption{Top: Pearson correlation between manual judgements of syntactic similarity and the measures ROUGE-1, ROUGE-L, BLEU, and their average. Bottom: Pearson correlation between manual judgements of semantic similarity and the measures Word Mover Score (WMS), BERT-Score (BERT), Sentence Transformer (ST), and their average.}
\label{table-correlation}
\end{table}

\providecommand{\bsc}{\color{darkgray}}

\begin{table*}

\fontsize{7pt}{10pt}\selectfont
\centering
\setlength{\tabcolsep}{2pt}
\renewcommand{\arraystretch}{1.0}

\begin{tabular}{@{}l@{}c@{}rccc cccccccc cccccccc cc@{}}

\toprule
\textbf{Corpus} & \multicolumn{5}{c}{\textbf{Corpus statistics}} & \multicolumn{8}{c}{\textbf{Syntactic similarity}} & \multicolumn{8}{c}{\textbf{Semantic similarity}} & & \\

\cmidrule(r{\tabcolsep}r{2\tabcolsep}){2-6}
\cmidrule(r{\tabcolsep}r{2\tabcolsep}){7-14}
\cmidrule(r{\tabcolsep}r{2\tabcolsep}){15-22}

Subset & \kern-0.5em Type & \multicolumn{2}{c}{Number} & \multicolumn{2}{c}{Length\quad} & \multicolumn{2}{c}{ROUGE-1} & \multicolumn{2}{c}{ROUGE-L}& \multicolumn{2}{c}{BLEU} & \multicolumn{2}{c}{Avg.} & 
\multicolumn{2}{c}{WMS} & \multicolumn{2}{c}{BERT} & \multicolumn{2}{c}{ST} & \multicolumn{2}{c}{Avg.} & \multicolumn{2}{c}{$\Delta_{\text{sem}, \text{syn}}$} \\

\arrayrulecolor{lightgray}
\cmidrule(l{\tabcolsep}r{\tabcolsep}){3-4}
\cmidrule(l{\tabcolsep}r{\tabcolsep}){5-6}
\cmidrule(r{\tabcolsep}r{\tabcolsep}){7-8}
\cmidrule(r{\tabcolsep}r{\tabcolsep}){9-10}
\cmidrule(r{\tabcolsep}r{\tabcolsep}){11-12}
\cmidrule(r{\tabcolsep}r{\tabcolsep}){13-14}
\cmidrule(r{\tabcolsep}r{\tabcolsep}){15-16}
\cmidrule(r{\tabcolsep}r{\tabcolsep}){17-18}
\cmidrule(r{\tabcolsep}r{\tabcolsep}){19-20}
\cmidrule(r{\tabcolsep}r{\tabcolsep}){21-22}
\cmidrule(r{\tabcolsep}r{\tabcolsep}){23-24}
\arrayrulecolor{black}

ST $> 0.8$ & & \multicolumn{1}{r}{$n$} & $\%$ & $\mu$ &\multicolumn{1}{c}{\bsc $\sigma$} & $\mu$ & \bsc $\sigma$ & $\mu$ & \bsc $\sigma$ & $\mu$ & \bsc $\sigma$ & $\mu$ & \bsc $\sigma$ & $\mu$ & \bsc $\sigma$ & $\mu$ & \bsc $\sigma$ & $\mu$ & \bsc $\sigma$ & $\mu$ & \bsc $\sigma$ & $\mu$ & \bsc $\sigma$ \\
\midrule
\addlinespace
$\text{\webisWikiICP}_\text{gold}$                   & C    &    10,327 &   34\% &  17.97 & \bsc  9.97 &  0.74 & \bsc  0.18 &  0.71 & \bsc  0.20 &  0.56 &  \bsc  0.28 &  0.67 & \bsc  0.21 &  0.83 & \bsc  0.12 &  0.69 &  0.19 &  0.91 & \bsc  0.06 &  0.81 & \bsc  0.11 &  0.14 & \bsc  0.13 \\
$\text{\webisWikiICP}_\text{silver}$                   & C    &    105,475 &   29\% &  20.38 & \bsc  13.56 &  0.71 & \bsc  0.19 &  0.67 & \bsc  0.22 &  0.52 &  \bsc  0.28 &  0.63 & \bsc  0.22 &  0.63 & \bsc  0.20 &  0.81 &  0.12 &  0.90 & \bsc  0.06 &  0.78 & \bsc  0.12 &  0.15 & \bsc  0.12 \\[0.5em]
Flickr8k                                 & C    &     29,011 &   36\% &  11.65 & \bsc   3.70 &  0.53 & \bsc  0.16 &  0.48 & \bsc  0.17 & \bf 0.22 &  \bsc  0.14 &  0.41 & \bsc  0.14 &  0.59 & \bsc  0.13 &  0.73 &  0.10 &  0.86 & \bsc  0.05 &  0.73 & \bsc  0.08 &  \bf 0.32 & \bsc  0.09 \\
PASCAL                                   & C    &      3,329 &   33\% &   9.83 & \bsc   3.30 & \bf 0.51 & \bsc  0.16 &  0.47 & \bsc  0.17 &   \bf 0.22 &  \bsc  0.14 &  0.40 & \bsc  0.15 &  0.59 & \bsc  0.13 &  0.72 &  0.10 &  0.86 & \bsc  0.05 &  0.73 & \bsc  0.08 &  \bf 0.32 & \bsc  0.09 \\
MS-COCO                                  & C    &    392,248 &   32\% &  10.54 & \bsc   2.25 & \bf 0.51 & \bsc  0.16 & \bf 0.45 & \bsc  0.16 &  \bf 0.22 &  \bsc  0.13 & \bf 0.39 & \bsc  0.14 &  0.57 & \bsc  0.14 &  0.71 &  0.10 &  0.86 & \bsc  0.04 &  0.71 & \bsc  0.08 & \bf 0.32 & \bsc  0.09 \\
PAWS                                     & G  &     64,940 &   99\% &  21.36 & \bsc   5.47 &  0.94 & \bsc  0.03 &  0.79 & \bsc  0.12 &  0.69 &  \bsc  0.18 &  0.81 & \bsc  0.10 & \bf 0.82 & \bsc  0.10 & \bf 0.96 &  0.04 & \bf 0.97 & \bsc  0.03 & \bf 0.92 & \bsc  0.04 &  0.11 & \bsc  0.08 \\
ParaNMT-5m                               & G  &  2,607,580 &   49\% &  11.97 & \bsc   6.18 &  0.63 & \bsc  0.15 &  0.60 & \bsc  0.16 &  0.33 &  \bsc  0.19 &  0.52 & \bsc  0.16 &  0.60 & \bsc  0.14 &  0.75 &  0.17 &  0.87 & \bsc  0.05 &  0.74 & \bsc  0.09 &  0.22 & \bsc  0.13 \\
MSRPC                                    & H      &      3,720 &   64\% &  20.02 & \bsc   4.94 &  0.73 & \bsc  0.12 &  0.69 & \bsc  0.13 &  0.54 &  \bsc  0.20 &  0.65 & \bsc  0.14 &  0.72 & \bsc  0.11 &  0.82 &  0.08 &  0.90 & \bsc  0.05 &  0.82 & \bsc  0.07 &  0.16 & \bsc  0.10 \\
PPDB 2.0                                 & H      &    654,531 &   24\% &   4.53 & \bsc   0.69 &  0.64 & \bsc  0.18 &  0.63 & \bsc  0.18 &  0.32 &  \bsc  0.18 &  0.53 & \bsc  0.17 &  0.64 & \bsc  0.22 &  0.63 &  0.30 &  0.89 & \bsc  0.05 &  0.72 & \bsc  0.14 &  0.19 & \bsc  0.15 \\
TaPaCo                                   & H      &    117,447 &   52\% &   5.47 & \bsc   2.28 &  0.65 & \bsc  0.18 &  0.63 & \bsc  0.18 &  0.30 &  \bsc  0.16 &  0.53 & \bsc  0.16 &  0.78 & \bsc  0.14 &  0.79 &  0.21 &  0.91 & \bsc  0.06 &  0.83 & \bsc  0.10 &  0.30 & \bsc  0.13 \\

\bottomrule
\end{tabular}

\caption{Comparison of syntactic similarity (measured as the average of ROUGE-1, ROUGE-L, and BLEU) and paraphrase semantic similarity (measured as the average of BERTScore, Sentence Transformer (ST) and Word Mover Score (WMS)) of semantically similar examples (Sentence Transformer $>$ 0.8) from state-of-the-art paraphrase corpora, image caption corpora, and the new silver-quality \webisWikiICP. ``Type'' classifies the acquisition method in caption (C), automatically generated (G), and human-written (H) (e.g., acquired via crowd-sourcing).}
\label{table-corpora-syntax-semantic-diff}
\end{table*}

Figure~\ref{fig:distribution-manual-judgements} shows the distribution of semantic and syntactic similarities of the median rating from the annotation study. These figures show that the semantic and syntactic similarities of the examples from the three different datasets are uniquely distributed. Examples from the TaPaCo corpus are pairs with subtle differences at the syntactic level that ultimately result in minor semantic differences. In contrast, examples from the \webisWikiICP (and MSRPC to some extent) are more evenly distributed along the diagonal syn-0/sem-0, \ldots, syn-4/sem-4. While \webisWikiICP has examples from all semantic similarity levels, semantic similarity is higher among examples from the~MSRPC.

To justify the choice of the semantic and syntactic similarity measures, we compute Pearson correlation coefficients with manual similarity judgments. Table~\ref{table-correlation} shows high correlation values for syntactic and semantic similarity ratings, which renders our measures well suited for the given task. 

\paragraph{Quantitative Similarity Analysis}

In the paraphrase literature, authors usually require pairs of texts to have varying degrees of semantic similarity to consider them paraphrases of each other. The choice of a ``semantic threshold'' drastically affects the distribution of syntactic and semantic similarities of paraphrases within a dataset. To standardize the decision criteria for paraphrases, we manually set a semantic threshold that allows us to comparably evaluate paraphrases from each dataset. 

We choose the semantic similarity score from the Sentence Transformer as the decision criterion for the subset of paraphrases under consideration because it has the highest correlation with human ratings of semantic similarity. To find a reliable threshold, we analyze the distributions of Sentence Transformer scores of paraphrases and non-paraphrases within the MSRPC and found that a threshold of~0.8 correctly classifies paraphrases with a precision and recall of 82\% and 77\%, respectively.

Table~\ref{table-corpora-syntax-semantic-diff} shows the results of the comparison of syntactic and semantic similarities of paraphrase pairs with a high semantic similarity (ST~>~0.8) grouped by type which is either image caption~(C), automatically generated~(G), or human-written~(H). $f$~calculates the frequency of text pairs that exceeds the semantic threshold of~0.8 relative to all examples in the dataset while $\Delta_{\text{sem},\text{syn}}$ computes the average difference between the average semantic and syntactic similarity scores. 

The high syntactic and semantic similarities of the examples from the PAWS dataset are striking. PAWS was designed as a benchmark corpus for paraphrase detection and contains lexically similar text pairs with subtle semantic inconsistencies that are difficult to distinguish from actual paraphrases. The high ROUGE-1 value of 94\% with a low standard derivation of 0.03 translates to an almost identical wording. Hence, it is no surprise that the $\Delta_{\text{sem},\text{syn}}$ is the lowest among all datasets. 

From the gold and silver quality proportions of the \webisWikiICP, 34\% and 29\% of the examples exceed the semantic threshold, respectively. These rates are in line with those of the other caption corpora studied. Furthermore, we find that all caption datasets have similar overall syntactic and semantic similarities. However, the caption pairs from the \webisWikiICP tend to have higher syntactic similarity. We suspect that this observation is due to Wikipedia's caption reuse policy, which allows editors to use slightly modified existing captions rather than inventing a new text.  However, the sufficiently large $\Delta_{\text{sem}, \text{syn}}$ of the \webisWikiICP, which is on par with the commonly used MSRPC, and larger than that of the other caption corpora strongly suggests the suitability of image captions as a source of interesting paraphrases. 

\section{Conclusion and Future Work}

We propose a new approach to use image captions as a resource for paraphrasing. From Wikipedia, we extracted 30,237~caption pairs in gold quality, 229,877~in silver quality, and 656,560~in bronze quality for our new \webisWikiICP dataset. As part of our analysis, we found that many caption pairs are ``sophisticated'' paraphrases in the sense of being semantically similar but dissimilar at the lexical and syntactic levels. We have introduced a respective measure for assessing paraphrase sophistication based on semantic and syntactic similarity. The new measure correlates well with manual judgments, and we have shown that paraphrases from different sources have individual characteristics along the two similarity dimensions. Studying large sets of paraphrases with characteristics required by a specific task (e.g., sophisticated vs.\ only word order changed) will be useful for task-specific paraphrase recognition and generation approaches, and for natural language understanding as a whole. 

In future work, we plan to extend the mining process to larger web resources than the Wikipedia and to apply image analysis as a new equivalence criterion. Identifying highly similar but not identical images may help to identify even more caption-based paraphrase candidates.

\section*{Limitations}

Challenging for our proposed paraphrase acquisition approach are captions that describe the same image in very different contexts (e.g.,~maps). Such captions tend to focus on rather different aspects of the image (e.g.,~rivers vs.\ cities on a map) but since the captions often still share some commonalities, the semantic component of our proposed measure $\Delta_{\text{sem}, \text{syn}}$ might still classify the captions as similar. A ``deeper'' estimation of semantic similarity could help. With more research on image--text relationships, applying models such as CLIP~\cite{radford:2021} or Stable Diffusion~\cite{rombach:2022} to take image semantics into account might lead to improvements for our proposed paraphrase refinement measure.

\raggedright
\bibliographystyle{acl_natbib}
\bibliography{eacl23-lit}

\end{document}